# TrueLabel + Confusions: A Spectrum of Probabilistic Models in Analyzing Multiple Ratings

**Chao Liu**[†]  CRAIGLIU@TENCENT.COM
Tencent Inc, 38 Haidian St, Beijing, 100080, P. R. China

**Yi-Min Wang**  YMWANG@MICROSOFT.COM
Microsoft Research, 1 Microsoft Way, Redmond, WA 98052, USA

## Abstract

This paper revisits the problem of analyzing multiple ratings given by different judges. Different from previous work that focuses on distilling the true labels from noisy crowdsourcing ratings, we emphasize gaining diagnostic insights into our in-house well-trained judges. We generalize the well-known DAWIDSKENE model (Dawid & Skene, 1979) to a spectrum of probabilistic models under the same "TrueLabel + Confusion" paradigm, and show that our proposed hierarchical Bayesian model, called HYBRIDCONFUSION, consistently outperforms DAWIDSKENE on both synthetic and real-world data sets.

## 1. Motivation

Recent advent of online crowdsourcing services (*e.g.*, Amazon's Mechanical Turk) excites the machine learning community by making large amount of labeled data practical. Because of the low cost, crowdsourcing labels are usually given by anonymous lowly-paid non-experts, which sparks recent interest in recovering the true labels from noisy (or even malicious) labels (Whitehill et al., 2009; Welinder et al., 2010; Welinder & Perona, 2010; Raykar et al., 2009). In this paper, we study the same problem of analyzing multiple ratings, but in quite a different setting.

We are in a major Web search engine company, and train search rankers using human ratings on the relevance of tens of millions of (query, URL) pairs. As it is too risky to bet the search engine on crowdsourcing ratings, we have to carefully recruit human judges, rigorously train them, and continually monitor their quality during the work. Since these judges are well-trained and the rating task is considerably hard, the cost of each label becomes so expensive that even two ratings per (query, URL) pair are economically infeasible: note that we have millions of pairs to rate and the number keeps increasing. Instead, we hope that a human judge would function satisfactorily once qualified, and each (query, URL) pair is only rated by one judge.

A key component in controlling the judge quality is to blend a small set of "monitoring" (query, URL) pairs into judges' regular work without their knowledge. This set of (query, URL) pairs are rated by all judges under monitoring. By analyzing the multiple ratings on (query, URL) pairs in this monitoring set, we hope to correctly score the quality of each judge, and more importantly, to gain insights into what confusions each judge makes so that we could plan targeted tutoring and revisions to the rating guidelines. Therefore, different from previous work that focuses on recovering the true labels from low-cost noisy labels, we are more interested in diagnostic information about judge confusions. For this reason, this paper emphasizes on probabilistic models that use a confusion matrix to quantify the competency of each judge.

The DAWIDSKENE model (Dawid & Skene, 1979) is a good candidate for this purpose. It pioneers the "TrueLabel + Confusion" paradigm: each item has a true label, and the rating each judge assigns to it is the true label obfuscated through the judge's confusion matrix. Suppose the rating is on a $K$-level scale, a confusion matrix is a $K \times K$ matrix

[†] This work was done when the first author was employed by Microsoft Research at Redmond.

TrueLabel + Confusions: A Spectrum of Probabilistic Models in Analyzing Multiple Ratings

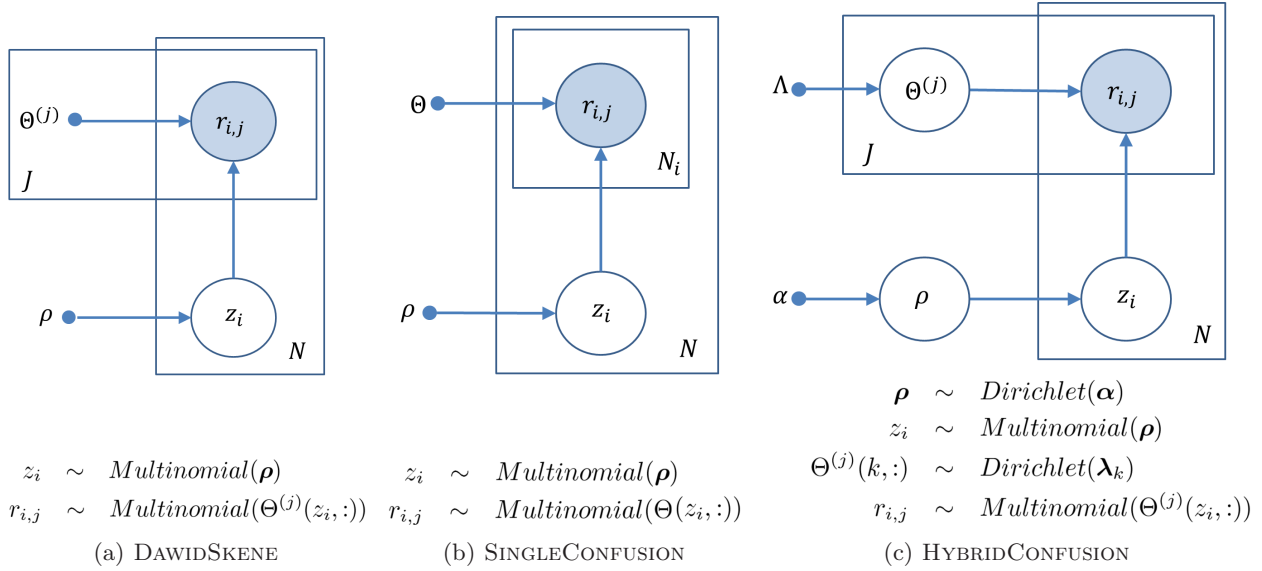

Figure 1. Graphical Model Representation of the Spectrum of Probabilistic Models

with the $(k,t)$ element being the probability that the judge would rate the item $t$ when the true label is $k$. Because each confusion matrix entails $K*(K-1)$ free parameters, DAWIDSKENE possesses at least $J*K*(K-1)$ free parameters when $J$ judges are involved. While this large number of free parameters renders big model capacity, it could also lead to overfitting easily, as will be seen in the experiments. For this reason, we propose a simplified model, called SINGLECONFUSION, which forces all judges to share the same confusion matrix. It significantly reduces the number of free dimensions, but unfortunately proves too rigid for real-world data, i.e., underfitting. As a tradeoff between the two, we further propose a hierarchical Bayesian model, called HYBRIDCONFUSION, which allows each judge to have her own confusion matrix, but at the same time regularizes these matrices through Bayesian shrinkage. Effectively, the three models form a spectrum of probabilistic models under the "TrueLabel + Confusion" paradigm. In summary, we make the following two contributions in this paper: (1) we study the problem of analyzing multiple ratings with an emphasis on the diagnostic aspect of different models, which complements previous work on recovering true labels from noisy ratings, (2) we generalize the well-known DAWIDSKENE model to a spectrum of probabilistic models under the same paradigm, and show the newly proposed model, HYBRIDCONFUSION, consistently outperforms DAWIDSKENE and SINGLECONFUSION on both synthetic and real-world data sets.

The rest of this paper is organized as follows. We elaborate on the spectrum of probabilistic models in Sections 2, and report on experiments on synthetic and real-world data in Sections 3 and 4, respectively. With related work discussed in Section 5, Section 6 concludes this study.

## 2. A Spectrum of Probabilistic Models

Suppose we have $N$ items rated by $J$ judges on a $K$-level metric. The metric can be either ordinal or categorical. Let $r_{i,j}$ be the rating of the $i$th item assigned by the $j$th judge: $r_{i,j} \in \{1,2,\cdots,K\}$ if the $j$th judge indeed rates the $i$th item and $r_{i,j}=0$ otherwise. We use $t_i \in \{1,2,\cdots,K\}$ to denote the true label of the $i$th item. The competency of the $j$th judge is modeled by a confusion matrix $\Theta^{(j)} \in R^{K \times K}$ with its $(k,t)$ element $\Theta^{(j)}_{k,t}$ being the probability that the $j$th judge will give a rating of $t$ when $t_i = k$. Collectively, we denote $\mathbf{r}_i = \{r_{i,j}\}_{j=1}^J$, $\mathbf{r} = \{\mathbf{r}_i\}_{i=1}^N$, $\mathbf{t} = \{t_i\}_{i=1}^N$, and $\Theta = \{\Theta^{(1)}, \Theta^{(2)}, \cdots, \Theta^{(J)}\}$, and use the hat notation ($\widehat{\phantom{x}}$) to denote the estimated value of the corresponding parameter.

The spectrum of probabilistic models are plotted in Figure 1. We start with a brief review of the DAWIDSKENE model (Figure 1(a)). It assumes that the true rating of each item is sampled from a multinomial distribution parameterized by $\rho$. Suppose the sampled true label is $k$, then the rating assigned by the $j$th judge is regarded as being sampled from another multinomial distribution parameterized by the

TrueLabel + Confusions: A Spectrum of Probabilistic Models in Analyzing Multiple Ratings

**Algorithm 1**: Inference for DAWIDSKENE

Input: Observed ratings **r**
Output: Estimated $\widehat{\Theta}, \widehat{\rho}$ and $\widehat{\mathbf{t}}$

**Initialize**:
$$\widehat{\rho}_k = 1/K, \widehat{\Theta}_{k,t}^{(j)} = \begin{cases} \lambda/(\lambda+K) & \text{if } k=t \\ 1/(\lambda+K) & \text{otherwise} \end{cases}$$

**Iterate until convergence**:
  **E-step**:
  $$Z_{i,k} = \frac{\rho_k \prod_{j=1}^{J} \mathbb{I}(r_{i,j} \neq 0) \Theta_{k,r_{i,j}}^{(j)}}{\sum_{k=1}^{K} \rho_k \prod_{j=1}^{J} \mathbb{I}(r_{i,j} \neq 0) \Theta_{k,r_{i,j}}^{(j)}}$$
  **M-step**:
  $$\widehat{\Theta}_{k,t}^{(j)} = \frac{\sum_{i=1}^{N} Z_{i,k} \mathbb{I}(r_{i,j}=t)}{\sum_{t=1}^{K} \sum_{i=1}^{N} Z_{i,k} \mathbb{I}(r_{i,j}=t)}$$
  $$\widehat{\rho}_k = \frac{\sum_{i=1}^{N} Z_{i,k}}{\sum_{k=1}^{K} \sum_{i=1}^{N} Z_{i,k}}$$

**Compute**:
$$\widehat{\mathbf{t}}_i = \underset{k=1,2,\cdots,K}{\arg\max} Z_{i,k}$$

---

**Algorithm 2**: Inference for SINGLECONFUSION

Input: Observed ratings **r**
Output: Estimated $\widehat{\Theta}, \widehat{\rho}$ and $\widehat{\mathbf{t}}$

**Initialize**:
$$\widehat{\rho}_k = 1/K, \widehat{\Theta}_{k,t} = \begin{cases} \lambda/(\lambda+K) & \text{if } k=t \\ 1/(\lambda+K) & \text{otherwise} \end{cases}$$

**Iterate until convergence**:
  **E-step**:
  $$Z_{i,k} = \frac{\rho_k \prod_{t=1}^{K} (\Theta_{k,t})^{\sum_{j=1}^{J} \mathbb{I}(r_{i,j}=t)}}{\sum_{k=1}^{K} \rho_k \prod_{t=1}^{K} (\Theta_{k,t})^{\sum_{j=1}^{J} \mathbb{I}(r_{i,j}=t)}}$$
  **M-step**:
  $$\widehat{\Theta}_{k,t} = \frac{\sum_{j=1}^{J} \sum_{i=1}^{N} Z_{i,k} \mathbb{I}(r_{i,j}=t)}{\sum_{t=1}^{K} \sum_{j=1}^{J} \sum_{i=1}^{N} Z_{i,k} \mathbb{I}(r_{i,j}=t)}$$
  $$\widehat{\rho}_k = \frac{\sum_{i=1}^{N} Z_{i,k}}{\sum_{k=1}^{K} \sum_{i=1}^{N} Z_{i,k}}$$

**Compute**:
$$\widehat{\mathbf{t}}_i = \underset{k=1,2,\cdots,K}{\arg\max} Z_{i,k}$$

---

$k$th row of the $j$th judge's confusion matrix, namely, $\Theta^{(j)}(k,:)$, using the Matlab notation. The goal of the inference is to recover the model parameters ($\Theta$ and $\rho$) and the true ratings **t**. To be self-contained and to compare with other models, the inference algorithm is reproduced in Algorithm 1 from (Dawid & Skene, 1979). The particular way of initialization in Algorithm 1 is to be consistent with HYBRIDCONFUSION, as to be discussed soon.

DAWIDSKENE model imposes no regularization on the individual confusion matrices, and hence possesses $(J * K + 1) * (K - 1)$ free parameters. In the first place, the large number of free parameters endows DAWIDSKENE with high model capacity, but in the second place, it means DAWIDSKENE could easily overfit and suffer from not having enough data to fit. This observation, as supported by experiments in Sections 3 and 4, prompts us to reduce the capacity by reducing the number of free parameters. We therefore propose the SINGLECONFUSION model, which forces all judges to have the same confusion matrix. Its graphical model and inference algorithm are presented in Figure 1(b) and Algorithm 2, respectively. SINGLECONFUSION effectively reduces the number of free parameters to $K^2 - 1$, but unfortunately, proves to be too rigid to model the variations across judges.

We can view DAWIDSKENE and SINGLECONFUSION as two extremes under the same "TrueLabel + Confusion" paradigm: one has too many parameters while the other has too few. We therefore propose the HYBRIDCONFUSION model, whose graphical model is depicted in Figure 1(c). HYBRIDCONFUSION makes tradeoffs between DAWIDSKENE and SINGLECONFUSION by allowing each judge to still have her own confusion matrix but at the same time regularizing them through Bayesian shrinkage. Explicitly, HYBRIDCONFUSION imposes that the $k$th row of all confusion matrices are sampled from a Dirichlet distribution parameterized by $\boldsymbol{\lambda}_k$. Explicitly, let

$$\Lambda = \begin{pmatrix} \lambda+1 & 1 & \cdots & 1 \\ 1 & \lambda+1 & \cdots & 1 \\ \vdots & \vdots & \ddots & \vdots \\ 1 & 1 & \cdots & \lambda+1 \end{pmatrix} = \begin{pmatrix} \boldsymbol{\lambda}_1 \\ \boldsymbol{\lambda}_2 \\ \vdots \\ \boldsymbol{\lambda}_K \end{pmatrix},$$

and it explains the choice of initialization in Algorithm 1 and 2. To complete the Bayesian model, a Dirichlet prior is imposed on $\rho$ as well with parameter $\boldsymbol{\alpha}$. We used Markov Chain Monte Carlo (Gibbs sampling in particular) to perform the inference on HYBRIDCONFUSION using the BUGS software package (Lunn et al., 2000). We run 3 Gibbs sampler with 1000 burn-in and obtain 100 examples with a thinning interval of 10 in all experiments. After collecting the samples, we take the mode rating from the true label samples as the recovered true label, and the mean values of samples about $\rho$ and $\Theta$ as the estimates $\widehat{\rho}$ and $\widehat{\Theta}$. As the $K$ labels are mutually exchangeable, all of the three models need to deal with the un-identifiability issue, and we tackle this using a method similar to (Stephens, 1999). In all the exper-

|   | A   | B   | C   |
|---|-----|-----|-----|
| A | 0.8 | 0   | 0.2 |
| B | 0.1 | 0.8 | 0.1 |
| C | 0.1 | 0   | 0.9 |

(a) $\Theta^{(1)}$

|   | A   | B   | C   |
|---|-----|-----|-----|
| A | 0.7 | 0.2 | 0.1 |
| B | 0.1 | 0.7 | 0.2 |
| C | 0.1 | 0.1 | 0.8 |

(b) $\Theta^{(2)}$

|   | A   | B   | C   |
|---|-----|-----|-----|
| A | 0.6 | 0.3 | 0.1 |
| B | 0.1 | 0.5 | 0.4 |
| C | 0.1 | 0.4 | 0.5 |

(c) $\Theta^{(3)}$

*Figure 2.* Confusion Matrices in the Simulation

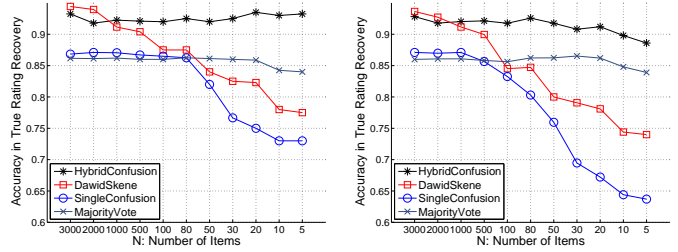

(a) Recovery Rate ($\lambda = 3$)  (b) Recovery Rate ($\lambda = 10$)

*Figure 3.* Accuracy in Recovering the True Label with Different $\lambda$'s

iments, $\boldsymbol{\alpha} = [1, 1, \cdots, 1]$ and $\lambda = 3$ unless otherwise noted.

The usage of $\lambda = 3$ imposes a moderate prior that implies judges are more likely to rate correctly in general. Depending on the applications, one could put a stronger prior by using a bigger $\lambda$ or impose confusion patterns based on prior knowledge (*e.g.*, a bigger value for $\Lambda_{1,3}$ if judges tend to misjudge "1" as "3"). In the case of ordinal rating, one could even put on a diagonal-decaying prior (*i.e.*, $\Lambda_{i,j_1} > \Lambda_{i,j_2}$ if $|j_1 - i| < |j_2 - i|$), indicating the belief that a judge is more likely to confuse adjacent levels than nonadjacent ones. In short, HYBRIDCONFUSION provides a flexible way to encode different prior knowledge, but the best setting, as always, depends on the application and prior knowledge.

## 3. Experiments on Synthetic Data

In this section, we use synthetic data to examine the accuracy of different models in recovering the ground truth, *i.e.*, estimating the true model parameters and labels. We synthesize the data using the DAWIDSKENE model for three judges ($J = 3$) on a three-level metric ($K = 3$). For convenience, we denote the three levels by "A", "B", and "C", whose prior probabilities are assumed to be 0.05, 0.15 and 0.8, respectively. The three confusion matrices $\Theta$ are shown in Figure 2. As can be seen, the three matrices are not chosen to favor any models, and in fact, $\Theta^{(1)}$ even disfavors HYBRIDCONFUSION because HYBRIDCONFUSION never estimates any cells to be 0.

Throughout the experiments in this section, we examine the accuracy of recovering the ground truth for different models by varying the number of items. Intuitively, the more rated items, the more accurately these models would recover the ground truth. The number of items is varied from 3000 down to 5, which renders a complete view of the model efficacy w.r.t. the data size. The data is synthesized in each run, and all the reported numbers are the average across 100 runs. We also include the majority voting (denoted by MAJORITYVOTE) algorithm in the comparison. MAJORITYVOTE takes the mode rating as the true rating, and breaks ties randomly when there are multiple modes. Once the true label is determined, $\widehat{\boldsymbol{\rho}}$ and $\widehat{\Theta}$ are obtained through simple counting.

### 3.1. Experimental Results

Figure 3(a) plots the accuracy in recovering the true labels when the number of items decreases from 3000 to 5, using the default parameter $\lambda = 3$. First, we observe that since the data is synthesized through DAWIDSKENE, the best recovery is indeed achieved by DAWIDSKENE, provided abundant data is available. But when data becomes smaller, DAWIDSKENE quickly deteriorates, as a result of overfitting. On the other hand, we see that HYBRIDCONFUSION is consistently above 0.9, only being slightly weaker than DAWIDSKENE when $N = 2000, 3000$, and significantly outperforming DAWIDSKENE otherwise. This consistent performance should be attributed to the Bayesian shrinkage. Third, we see MAJORITYVOTE is pretty flat, and effectively saturates after $N \geq 20$. Finally, SINGLECONFUSION is the weakest and twisted with MAJORITYVOTE when $N \geq 50$. The inferior performance of SINGLECONFUSION is likely due to its strict constraint to have all judges share the same confusion matrix whereas in fact they are quite different (see Figure 2). Figure 3(b) presents the same experiments with $\lambda = 10$, which deliver similar messages.

The accuracy of recovering $\boldsymbol{\rho}$ by HYBRIDCONFUSION, DAWIDSKENE, and SINGLECONFUSION with varying numbers of items are plotted in Figure 4. Each bar represents the recovered $\hat{\rho}_1$, $\hat{\rho}_2$ and $\hat{\rho}_3$ for a given algorithm when a certain of number of items are given. The recovered $\boldsymbol{\rho}$ by MAJORITYVOTE is consistently around (0.1, 0.15, 0.75) regardless of the number of items, and hence not plotted in Figure 4. We see that both HYBRIDCONFUSION and DAWIDSKENE can exactly recover the true $\boldsymbol{\rho}$ with enough data whereas SINGLECONFUSION misses the target even when 3000 items are provided.

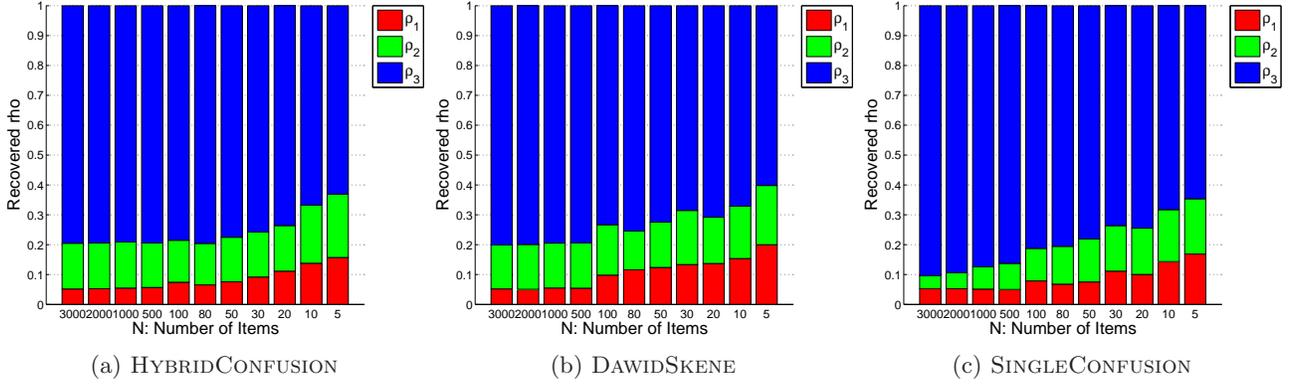

Figure 4. Accuracy in Recovering $\rho$ by Different Models

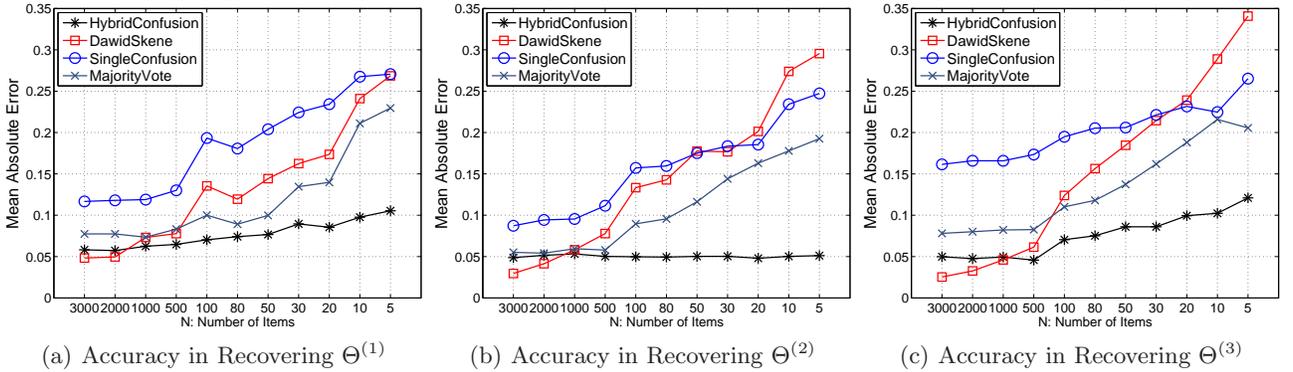

Figure 5. Mean Absolute Error in Recovering $\Theta$

Finally, Figure 5 plots the accuracy in recovering the confusion matrices, as measured by the Mean Absolute Error (MAE),

$$MAE(\widehat{\Theta}^{(i)}, \Theta^{(i)}) = \frac{1}{K^2} \sum_{k=1}^{K} \sum_{t=1}^{K} |\widehat{\Theta}^{(i)}_{k,t} - \Theta^{(i)}_{k,t}|.$$

Figure 5(a) plots the MAE of different models in recovering $\Theta^{(1)}$ when the number of items varies. Again, we have seen that DAWIDSKENE gives the best recovery, but only when enough ratings are available, and it clearly deteriorates with fewer and fewer data. HYBRIDCONFUSION, on the other hand, is very accurate, and barely decays when $N$ gets smaller. SINGLECONFUSION remains the worst performing model but MAJORITYVOTE appears very competitive. The same trend on relative performance of different models is confirmed in Figures 5(b) and 5(c) as well.

As a short conclusion, we observe that HYBRIDCONFUSION is comparable to DAWIDSKENE when abundant data is available even if the data is synthesized by DAWIDSKENE. DAWIDSKENE quickly deteriorates when data becomes scarce, but HYBRIDCONFUSION remains very accurate in recovering the ground truth, which clearly demonstrates the efficacy of Bayesian shrinkage. But, to be fair, DAWIDSKENE is about 4 times faster than HYBRIDCONFUSION.

## 4. Experiments on Real-World Data

In this section, we report on the experimental results on a real-world data set that motivates the study. As explained in Section 1, we use a set of "monitoring" (query, URL) pairs to gauge judge quality. For each (query, URL) pair, a judge is expected to assign it into one of the five categories *Bad, Fair, Good, Excellent, Perfect* (denoted by 1 to 5, respectively), based on her assessment of the relevance according to a set of written guidelines. This "monitoring" set is chosen to be "hard" to maximally differentiate judge qualities. A "super-judge", who supposedly best understands the judgment guidelines, gives the gold rating to each pair in the set. The quality of each judge is then determined based on the deviation from gold ratings. Because only a few super-judges are available, quality calibra-

TrueLabel + Confusions: A Spectrum of Probabilistic Models in Analyzing Multiple Ratings

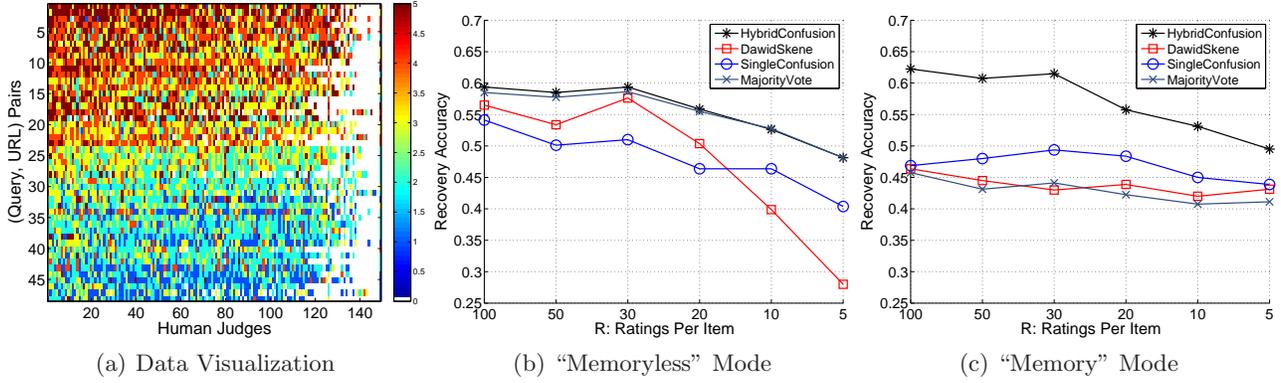

(a) Data Visualization  (b) "Memoryless" Mode  (c) "Memory" Mode

Figure 6. Experiments on Real-World Data Set

tion based on existing gold labels usually suffers from data scarcity. For example, a lot of cells in the confusion matrices would be zero. Therefore, in this section, we explore to what extent each model would recover the gold rating based on individual ratings from regular judges, and furthermore whether the estimated parameters by each model could help further lift the prediction accuracy.

One monitoring set consists of 6008 ratings from 148 judges on 48 (query, url) pairs, as visualized in Figure 6(a). Each row corresponds to a distinct (query, url) pair, and each column to a distinct human judge. For eligibility, (query, url) pairs and judges are sorted based on the average rating across judges and items, respectively. The gold labels are listed in the last column. There are only 6 colors of interest in the colorbar: blue for 1 (*Bad*) to red for 5 (*Perfect*) with white for 0 denoting unrated item by judges. As can be seen, some judges only rated a few (query, URL) pairs, and they are generally new judges hired into the system. They are not excluded in order to keep the fidelity to real-world data.

We report experiment results using the average across 100 runs. In each run, we first sample $R$ ratings for each item, and then randomly partition the 48 (query, URL) pairs into training and testing sets with a ratio of 2:1. We compare the accuracy of predicting the gold label on the testing set with different models. In order to see if any models could recover the unknown true parameters to some extent from the training set, we test the prediction accuracy in two different modes. The first is the "Memoryless" mode where the true label is predicted solely based on the testing set, and the second is the "Memory" mode where each (query, URL) pair is predicted using the estimated parameters obtained from the training set. Explicitly, the "Memory" mode computes the posterior of the rating based on $\widehat{\Theta}, \widehat{\rho}$ and $\mathbf{r_i}$ using

$$p(z_i = k|\mathbf{r_i}; \widehat{\Theta}, \widehat{\rho}) = \frac{\widehat{\rho}_k \prod_{j=1}^J \mathbb{I}(r_{i,j} \neq 0)\widehat{\Theta}_{k,r_{i,j}}^{(j)}}{\sum_{k=1}^K \widehat{\rho}_k \prod_{j=1}^J \mathbb{I}(r_{i,j} \neq 0)\widehat{\Theta}_{k,r_{i,j}}^{(j)}},$$

and assigns the $i$th item to the most probable rating.

The "Memoryless" model corresponds to the cold-start scenario where no knowledge about the judges is available, whereas the "Memory" mode mimics the case where the confusion matrices and label distributions are known a priori from previous data. However, in neither mode is the gold label of the training set used, and the gold label of testing data is merely used in measuring the prediction accuracy. The goal of this study aims to comparing the recovery accuracy of the true label using different models, and to what extent the estimated parameters could help; the goal is not to build a state-of-the-art prediction model for true labels.

Figures 6(b) and 6(c) plot the prediction accuracy w.r.t. the number of ratings per item in the two modes, respectively. In the "Memoryless" mode (Figure 6(b)), all models deteriorate as fewer ratings are available to each item, and DAWIDSKENE deteriorates the most, yet another piece of evidence of overfitting. In contrast, HYBRIDCONFUSION performs very well and does not deteriorate much even when $R = 5$. Surprisingly, MAJORITYVOTE is very close to HYBRIDCONFUSION, and the performance gap is no longer as wide as that shown in Figure 3. A likely reason for the strong performance of MAJORITYVOTE is that our judges are all well-trained and understand that their quality is continually monitored; hence their consensus is usually much stronger than that between crowdsourcing judges. Nevertheless, we still see HYBRIDCONFUSION beats MAJORITYVOTE with a small but consistent margin with 30 or more ratings per item ($p$-value $< 0.05$ using one-sided Fisher Sign Test).

TrueLabel + Confusions: A Spectrum of Probabilistic Models in Analyzing Multiple Ratings

Figure 6(c) compares the performance in the "Memory" mode, where we see that HYBRIDCONFUSION significantly beats all other three models and is visibly better than HYBRIDCONFUSION in the "Memoryless" mode. This suggests that HYBRIDCONFUSION indeed recovers the true parameters to some extent, which helps lift the prediction accuracy. In contrast, we see the accuracy of MAJORITYVOTE drops significantly from the "Memoryless" mode, indicating the inferior quality of the estimated parameters due to data sparsity.

Finally, we peek into the estimated confusion matrices by DAWIDSKENE and HYBRIDCONFUSION to understand what the estimates look like. Figure 7 plots the average confusion matrices for DAWIDSKENE and HYBRIDCONFUSION, as averaged across all judges over the 100 runs with $R = 5$. At the first glimpse, we see the confusion matrix from DAWIDSKENE is much more chaotic than that from HYBRIDCONFUSION, which visually demonstrates the efficacy of Bayesian shrinkage in HYBRIDCONFUSION. Within each cell, we also mark out the probability value with the standard deviation in the parenthesis. Clearly, we see that confusion matrices from DAWIDSKENE are more divergent from each other (bigger standard deviations as shown in Figure 7(a)). In contrast, the confusion matrix from HYBRIDCONFUSION (Figure 7(b)) exhibits natural diagonal-decaying phenomena, and the standard deviations are generally much smaller. Now that our judges are well-trained, we believe the latter is closer to the truth than the former, although we have no way to solicit the ground truth.

## 5. Related Work

The need of large amount of labeled data, as inherent in many machine learning algorithms and applications, cultivates the advent of online crowdsourcing services (*e.g.*, Amazon's MechanicTurk). Readers interested in a detailed survey in this area are referred to (Ipeirotis & Paritosh, 2011). While the low cost of crowdsourcing renders multiple labels on numerous items practically feasible, it also calls for principled approaches to distilling true labels from the less-than-expert ratings, among other issues (Sheng et al., 2008). Because of the importance of this problem, recent years have seen increasing interests in this problem, *e.g.*, (Whitehill et al., 2009; Welinder et al., 2010; Welinder & Perona, 2010; Raykar et al., 2009). Specifically, Whitehill et al. (2009) models the probability that a judge would rate an item correctly as a logistic function of the product between the quality of the judge and the difficulty of the item. This leads to a model that not only recovers the true label, but estimates the judge quality (represented by a single number) and item difficulty at the same time. But since the model deals with the probability that a judge hits the right rating, it does not provide detailed diagnostic information about judge confusions. (Welinder et al., 2010; Welinder & Perona, 2010) later generalizes (Whitehill et al., 2009) by introducing a high-dimensional concept of item difficulty, and shows a small but consistent improvement, but again it fails to provide the confusion matrices. In this paper, we focus on models under the "TrueLabel + Confusion" paradigm for diagnostic insights into judge confusion.

In addition to the different focus, the judges in our setting are also different from previous work: our in-house judges are well trained, decently paid, and benign in general, which directly contrasts to the anonymous non-expert or even malicious judges in crowdsourcing services. This entails two consequences: first, we do not worry about malicious judges, and second, MAJORITYVOTE becomes much more competitive, although it is still inferior to our proposed model.

The spectrum of models as presented here are not restricted to ratings from human judges, and in fact they can be effectively used to combine any ratings from any sources, be it human judges or predictive models. Previously, Ghahramani and Kim presented some preliminary results using a model similar to HYBRIDCONFUSION (Ghahramani & chul Kim, 2003), but failed to examine how the performance varies when the numbers of items and judges change. This paper fills in the gap, and to the best of our knowledge, this is the first piece of work generalizing DAWIDSKENE to a spectrum of models with a comparative study of their pros and cons. Finally, another piece of work worth mentioning is (Raykar et al., 2009), which performs supervised learning with the true labels recovered as a by-product. The method is shown superior to the conventional two-stage alternative (*i.e.*, training models after recovering the true labels, as practiced by (Smyth et al., 1994)). Their focus is on building accurate predictive models rather than diagnosing judge qualities.

## 6. Conclusion

In this paper, we generalize the DAWIDSKENE model into a spectrum of probabilistic models under the "TrueLabel+Confusion" paradigm. Our proposed models, SINGLECONFUSION and HYBRIDCONFUSION, complement the well-known DAWIDSKENE model to overcome its overfitting drawbacks. We study their pros and cons using both synthetic and real-

|        | Bad  | Fair | Good | Excellent | Perfect |
|--------|------|------|------|-----------|---------|
| Perfect | 0.03 (0.08) | 0.17 (0.19) | 0.15 (0.18) | 0.22 (0.21) | 0.43 (0.27) |
| Excellent | 0.01 (0.04) | 0.08 (0.12) | 0.20 (0.17) | 0.43 (0.22) | 0.28 (0.19) |
| Good | 0.02 (0.06) | 0.10 (0.12) | 0.44 (0.18) | 0.27 (0.17) | 0.17 (0.15) |
| Fair | 0.21 (0.12) | 0.44 (0.17) | 0.24 (0.15) | 0.08 (0.10) | 0.03 (0.06) |
| Bad | 0.42 (0.39) | 0.20 (0.31) | 0.21 (0.32) | 0.15 (0.29) | 0.01 (0.09) |

(a) Confusion Matrix: DAWIDSKENE

|        | Bad  | Fair | Good | Excellent | Perfect |
|--------|------|------|------|-----------|---------|
| Perfect | 0.09 (0.02) | 0.13 (0.05) | 0.14 (0.05) | 0.16 (0.06) | 0.48 (0.08) |
| Excellent | 0.08 (0.02) | 0.11 (0.05) | 0.16 (0.08) | 0.47 (0.09) | 0.18 (0.08) |
| Good | 0.07 (0.02) | 0.12 (0.05) | 0.48 (0.07) | 0.19 (0.08) | 0.13 (0.06) |
| Fair | 0.14 (0.06) | 0.49 (0.08) | 0.18 (0.08) | 0.10 (0.04) | 0.08 (0.02) |
| Bad | 0.49 (0.08) | 0.17 (0.07) | 0.14 (0.05) | 0.11 (0.03) | 0.09 (0.02) |

(b) Confusion Matrix: HYBRIDCONFUSION

Figure 7. A Peek into the Average Confusion Matrix

world data, and demonstrate the advantages of HYBRIDCONFUSION in various settings. In the future, we would optimize the judgement pipeline based on the recovered confusion matrices, *e.g.*, through targeted training and guideline revisions.

## Acknowledgments

We would like to thank John Platt for the helpful discussion, and the anonymous reviewers for the insightful and constructive comments.